\documentclass[letterpaper,11pt]{article}

\usepackage[margin=1in]{geometry}

\usepackage[style=numeric-comp,sorting=none,maxnames=1,maxbibnames=3]{biblatex}

\renewbibmacro{in:}{}
\DeclareFieldFormat[article]{title}{#1}
\DeclareFieldFormat[book]{title}{#1}
\DeclareFieldFormat[thesis]{title}{#1}
\DeclareFieldFormat[article]{volume}{#1}

\AtEveryBibitem{
  \clearfield{note}
  \clearfield{month}
  \clearfield{doi}
  \clearfield{url}
  \clearfield{isbn}
  \clearfield{issn}
  \clearfield{howpublished}
}

\newcommand{\parabf}[1]{\noindent\textbf{#1}}

\usepackage{times}
\usepackage{amsmath,amsthm,amsfonts}
\usepackage{xspace}
\usepackage{graphicx}

\theoremstyle{plain}

\usepackage{float}
\usepackage[english]{babel}
\usepackage{subfig}
\usepackage{xcolor}
\usepackage{colortbl}
\usepackage{pifont}
\usepackage[inline]{enumitem}
\usepackage{multirow}
\usepackage{fancybox} %
\usepackage[small, compact]{titlesec}
\usepackage[textfont={it}, font={small,bf}]{caption}
\usepackage[normalem]{ulem}
\usepackage{authblk}
\usepackage{comment}
\usepackage{booktabs}

\newcommand{\mosharaf}[1]{\textcolor{blue}{\{MC: #1\}}}

\newcommand{\jw}[1]{\textcolor[rgb]{0.25,0.65,0.1}{\{JW: #1\}}}
\newcommand{\yile}[1]{\textcolor{orange}{\{YG: #1\}}}
\renewcommand{\mosharaf}[1]{}
\renewcommand{\jw}[1]{}
\renewcommand{\yile}[1]{}

\usepackage{tikz}
\usetikzlibrary{positioning}

\theoremstyle{definition}

\newtheorem*{defi*}{Definition}

\usepackage{hyperref}
\hypersetup{
	colorlinks=true,      %
	linkcolor=blue,       %
	citecolor=magenta,    %
	filecolor=cyan,       %
	urlcolor=red          %
}

\usepackage{titling}
\date{}

\usepackage{listings}
\usepackage{courier}
\begin{document}

	\date{}
	
  \title{\Large \bf Toward Cross-Layer Energy Optimizations in AI Systems}

	\author{
		\rm Jae-Won Chung \qquad Nishil Talati \qquad Mosharaf Chowdhury
    \\
    \normalsize \texttt{\{jwnchung,talatin,mosharaf\}@umich.edu}
		\\
		\itshape{University of Michigan}
	 } %

	\pagestyle{plain}
	
  \pagenumbering{gobble} %
	
	\maketitle

  \vspace{-1.5cm}

  \begin{center}
    \textbf{Topics: System software and Architecture}
  \end{center}

	\section{Context}

The ``AI for Science, Energy, and Security'' report~\cite{carter2023advanced} from DOE outlines a significant focus on developing and optimizing AI workflows for a foundational impact on a broad range of DOE missions.
With the pervasive usage of artificial intelligence (AI) and machine learning (ML) tools and techniques, their energy efficiency is likely to become the gating factor toward adoption.
This is because generative AI (GenAI) models are massive energy hogs: for instance, training a 200-billion parameter large language model (LLM) at Amazon is estimated to have taken 11.9 GWh~\cite{hamilton2024constraint}, which is enough to power more than a thousand average U.S. households for a year~\cite{us-household}. %
Inference consumes even more energy, because a model trained once serve millions.
Given this scale, high energy efficiency is key to addressing the power delivery problem of constructing and operating new supercomputers and datacenters specialized for AI workloads~\cite{cbre2024,mckinsey2023}.

	\section{Challenges}

\parabf{Optimization silos.}
Although energy optimization is well explored in the hardware community, it is often done in isolation from the software stack.
However, the emerging power bottleneck of AI clusters show that energy-efficiency gains from hardware alone are insufficient to sustain the growing demand for compute.
Indeed, recent works have shown software design can significantly impact AI energy consumption~\cite{zeus-nsdi23,perseus-sosp24}.

\parabf{Lack of a comprehensive model.}
Despite multi-generation improvements in both hardware (e.g., CPUs, GPUs) and software (e.g., CUDA, NCCL, PyTorch) efficiency, the community still lacks a comprehensive framework that closely models complex hardware/software interactions and identifies sources of inefficiencies in an end-to-end system.
This is especially crucial for multi-phase GenAI workloads, where compute and IO can be performance and energy bottlenecks in different ways in each phase of computation.

\parabf{Limited cross-layer understanding of application metrics.}
Optimizing system and infrastructure software is not enough either. 
In order to tease out energy efficiency opportunities, we need to precisely understand application-specific metrics and exploit them in low-level optimizations.
For instance, an AI serving system may have latency deadlines, where users do not care as long as requests complete before their deadlines.
With application-level deadlines exposed to the underlying hardware, the hardware could run at a slower speed to reduce energy consumption while still maintaining the deadline.

\parabf{Missing cross-layer control.}
Merely identifying cross-layer components and corresponding knobs that impact application metrics is not enough; we need an efficient method for searching through the combinatorial space of software and hardware knobs and locating the optimal combination.
This can be challenging depending on factors like algorithmic hardness, large search space size (including hardware choices and compute placement), or solver latency requirements.
For instance, choosing the minimum-energy batch size during training is already non-trivial.
This is because the number of training steps needed for the model to reach the target accuracy is both stochastic and hard to estimate before actually completing training.

\parabf{Maintaining generality.}
The same software can run on diverse hardware, and the same hardware can execute different types of software.
Especially for AI workloads, we expect increasingly heterogeneous accelerators (e.g., NVIDIA or AMD GPUs, custom accelerators like TPUs and Inferentia).
The software and hardware optimizations we make should not only be general enough to avoid coupling certain software and hardware, but also specific enough to reap most of the potential energy-efficiency gains.

	\section{Opportunity}

Generality cannot be attained without it being part of the core design.
To that end, we believe that we need to define a narrow interface to exchange the necessary and sufficient information between the hardware and software layers -- something akin to the ``narrow waist'' design in other systems (e.g., IP layer in networking, LLVM Intermediate Representation in compilers) that sits in between and guides the design of both.

As the narrow waist, a possibility we propose is the \emph{Pareto frontier} (tradeoff curve) of the time and energy consumption of a sequence of ML computations.
The frontier contains a set of Pareto-efficient (time, energy) points a sequence of ML computations can consume for execution, and each point on the frontier is induced by running the computation with a specific combination of knobs in both software (e.g., batch size for ML training and inference, computation/memory intensity of kernel, usage of kernel fusion) and hardware (e.g., GPU power limit, frequency, model/microarchitecture) layers.
The time--energy Pareto frontier has been shown to universally exist for ML computations across multiple generations of GPUs~\cite{zeus-nsdi23,perseus-sosp24}.

Under the narrow waist design, efficiency advancements of software and hardware layers can be \emph{decoupled}.
The software layer can assume that the underlying hardware provides some time--energy Pareto frontier and create optimization algorithms that make use of them.
For instance, when an ML serving system knows how much slack time a request has, it can select the point on the Pareto frontier that precisely slows down computation until the deadline, thereby extracting maximum energy-efficiency from the underlying hardware without missing the deadline.
More challenging time and energy allocation problems will arise in more complex scenarios, including large model training and DAG-based workflows.

Conversely, hardware can conveniently assume that the software layer has algorithms to make use of any time--energy Pareto frontier.
Then, new hardware architectures that push the time--energy Pareto frontier further or provide a wider dynamic range of time and energy can be developed.
Furthermore, new hardware control knobs that provide better time--energy Pareto frontiers can be created, and the software layer can easily utilize them as long as the hardware exposes control to the software layer.

Finally, the time--energy Pareto frontier can be used as a basic building block of modeling system dynamics and their end-to-end energy efficiency, enabling a wide range of what-if simulations.
For accurate estimations of time and energy, there is an opportunity to develop more detailed models that consider details like kernel launch overheads and data transfer/IO costs. 
Further, enhancing data points on this frontier with hardware-level details (e.g., compute and memory resource utilization) can provide users with valuable insights. 
Altogether, not only this approach allows for the optimization of existing infrastructures, but it also facilitates path finding for developing next-generation energy-efficient software/hardware AI stacks.

  \section{Timeliness and Impact}

Generative AI adoption and its energy consumption are skyrocketing with broad implications. %
First, energy-intensive AI workloads inflate operational expenses and carbon offsetting costs for entities with Net Zero commitments.
Second, they have made power delivery one of the primary challenges in building new AI clusters today.
Finally, this can hinder deploying AI services to places without high-capacity, stable electricity grids.
Only a coordinated software-hardware effort can holistically address these problems.

It is high time we invest in cross-layer energy efficiency optimizations for AI systems.

  \label{EndOfPaper}

  \printbibliography
	\clearpage

\end{document}